\documentclass[sigconf]{acmart}
\usepackage{enumitem}
\usepackage{makecell}
\usepackage{multirow}
\usepackage{float}
\usepackage{subcaption}
\usepackage{verbatim}
\usepackage{colortbl}

\usepackage{xcolor}

\definecolor{darkgreen}{RGB}{0,100,0}

\usepackage{mdframed} 

\usepackage{tcolorbox}
\usepackage{multirow}
\tcbuselibrary{breakable,skins}

\tcbset{
  promptbox/.style={
    enhanced,
    breakable,
    colback=white,
    colframe=black,
    boxrule=0.4pt,
    arc=0pt,
    boxsep=2pt,
    left=4pt,
    right=4pt,
    top=4pt,
    bottom=4pt,
    before skip=6pt,
    after skip=6pt,
    fontupper=\ttfamily\footnotesize
  }
}

\AtBeginDocument{%
  }

\copyrightyear{2026}
\acmYear{2026}
\setcopyright{cc}
\setcctype{by}
\acmConference[CIKM '26]{Proceedings of the 35th ACM International Conference on Information and Knowledge Management}{November 07--11, 2026}{Rome, Italy}
\acmBooktitle{Proceedings of the 35th ACM International Conference on Information and Knowledge Management (CIKM '26), November 07--11, 2026, Rome, Italy}
\acmDOI{10.1145/3799682.3840033}
\acmISBN{979-8-4007-2539-5/2026/11}

\begin{document}

\title{
Judging a Review by its Cover: A Reliability Analysis of LLM-based Peer Review Evaluation Metrics}

\author{Shakiba Amirshahi}
\orcid{https://orcid.org/0009-0009-1524-8643}
\affiliation{%
  \institution{Reviewerly}
  \city{Toronto}
  \state{Ontario}
  \country{Canada}
}

\author{Sajad Ebrahimi}
\authornote{Corresponding author. Email: \texttt{s.ebrahimi@utoronto.ca}}
\orcid{https://orcid.org/0009-0003-1630-3938}
\affiliation{%
  \institution{Reviewerly}
  \city{Toronto}
  \state{Ontario}
  \country{Canada}
}

\author{Hai Son Le}
\orcid{https://orcid.org/0009-0003-2240-0451}
\affiliation{%
  \institution{Reviewerly}
  \city{Toronto}
  \state{Ontario}
  \country{Canada}
}

\author{Negar Arabzadeh}
\orcid{https://orcid.org/0000-0002-4411-7089}
\affiliation{%
  \institution{University of California, Berkeley}
  \city{Berkeley}
  \state{California}
  \country{United States}
}
\affiliation{%
  \institution{Reviewerly}
  \city{Toronto}
  \state{Ontario}
  \country{Canada}
}

\author{Ebrahim Bagheri}
\orcid{https://orcid.org/0000-0002-5148-6237}
\affiliation{%
  \institution{University of Toronto}
  \city{Toronto}
  \state{Ontario}
  \country{Canada}
}
\affiliation{%
  \institution{Reviewerly}
  \city{Toronto}
  \state{Ontario}
  \country{Canada}
}

\renewcommand{\shortauthors}{Shakiba Amirshahi, Sajad Ebrahimi, Hai Son Le, Negar Arabzadeh, \& Ebrahim Bagheri}

\begin{abstract}
Peer-review evaluation is increasingly being automated with LLM-as-a-judge metrics, but this creates a measurement risk. A review may receive a high score because it is fluent, organized, and polished, rather than because it provides a strong evaluation of the paper. This risk is especially important in AI-assisted reviewing, where reviewers may use LLMs to improve clarity or presentation while preserving the underlying judgments. We propose a statistical framework for testing whether peer-review evaluation metrics capture substantive review quality beyond surface-level linguistic form. The framework compares original human reviews with faithful LLM rewrites that preserve the same evaluative content while changing wording and presentation. Using a dataset comprising 4,044 meaning-preserving rewrites derived from 674 human reviews from ICLR and NeurIPS, we evaluate 29 content-oriented peer-review evaluation metrics drawn from four prior works through complementary tests of \emph{surface sensitivity} and \emph{robustness}. Although these metrics are intended to capture review properties beyond surface-level, writing-dependent characteristics, we find that sensitivity to rewriting is widespread. Under our primary analysis, 23 metrics assign significantly different scores to reviews whose evaluative content is preserved, while only six satisfy our robustness criterion. The patterns are largely consistent across two LLM judge models, suggesting that the issue is not specific to a single judge. These findings show that many peer-review evaluation metrics partially conflate review quality with linguistic presentation, and indicate that robustness to meaning-preserving rewriting should be validated before such metrics are used to compare human-written, AI-assisted, and AI-generated reviews. All evaluation results, prompts, and implementation code are publicly available in our GitHub repository: {\url{https://github.com/Reviewerly-Inc/llm_judge_reliability_analysis}}.
\end{abstract}



\maketitle

\section{Introduction}

Peer review plays a central role in scientific publishing, shaping publication decisions, funding outcomes, and academic careers~\cite{kelly2014peer,tennant2017multi,horbach2018changing, ebrahimi2025exharmony}. Yet this process is increasingly strained by growing submission volumes, which place substantial pressure on reviewers, editors, and conference organizers. At the same time, large language models (LLMs) are beginning to change how reviews are written. LLMs are now used both to assist reviewers and to generate review text~\cite{liang2024monitoring,latona2024ai,liu2023reviewergpt}. This shift creates a parallel need for scalable methods that can evaluate peer-review quality without relying exclusively on costly human assessment~\cite{ebrahimi2026peeriscope,arabzadeh2026doxa, arabzadeh2025building}.

Peer-review-specific LLM-as-a-judge metrics have become a prominent approach for automated peer review evaluation, scoring reviews along dimensions such as analytical depth~\cite{revieweval}, structured review component coverage~\cite{remor}, holistic review quality~\cite{rottenreviews}, and human-anchored comparative judgment~\cite{scholarpeer}, with recent benchmarks such as PRISM~\cite{loc2026prism} and PRAIB~\cite{farganus2026praib} further expanding evaluation criteria. These developments make 
peer review evaluation more scalable, but they also raise a basic 
measurement question. A review can be well written without being 
a high-quality review, fluent, organized, and polished, while 
offering shallow analysis, weak evidence, or unhelpful 
recommendations, or containing a substantive and useful evaluation 
while being less polished in presentation. For LLM-as-a-judge metrics to support reliable peer review evaluation, they must distinguish the \textit{quality of the review as an evaluation} from the \textit{quality of the review as a piece of writing}~\cite{sizo2025defining}. Scores for substantive properties such as actionability, constructiveness, and depth of technical analysis should not change merely because the same review is expressed in a more polished style.

This paper studies a central reliability problem in LLM-based peer-review evaluation. We ask whether metrics intended to measure substantive review quality instead reward surface form. This is especially important in AI-assisted reviewing, where an LLM may improve clarity, fluency, or organization while preserving the reviewer’s underlying judgments~\cite{arabzadeh2026can}. If a metric assigns different scores to the original and rewritten versions of the same review, it may be responding to linguistic presentation rather than evaluative content, making polished reviews appear better even when their substantive assessment has not changed.

The concern is consistent with broader findings on LLM-based 
evaluation, including prompt sensitivity, position bias, 
judge-model disagreement, and preferences for particular writing 
styles~\cite{zheng2023judging,saito2023verbosity,wang2024large,
panickssery2024llm}, and connects to work on language invariance, 
where meaning-preserving transformations should not change 
properties intended to remain constant under the surface 
variation~\cite{bianchi2022language}. Similar questions have been 
studied in machine translation and summarization, where automatic 
metrics are expected to remain reliable under meaning-preserving 
transformations~\cite{marie2021scientific,
kocmi2021ship}. 
Despite the growing use of LLM-as-a-judge metrics for 
peer-review evaluation, their reliability has not 
been systematically examined to date.

We study this problem from two distinct complementary perspectives. \emph{Surface sensitivity} asks whether meaning-preserving rewriting induces a statistically detectable change in metric scores, identifying metrics that systematically respond to changes in wording or presentation despite preserved evaluative content. \emph{Robustness} asks whether any score change is small enough to be practically negligible~\cite{lakens2017equivalence}. 
These two perspectives capture different forms of metric reliability. 
\emph{Surface sensitivity} identifies whether rewriting changes a metric's scores in a systematic way, whereas \emph{robustness} determines whether the size of that change is small enough to be acceptable for practical evaluation. 
A metric can therefore be sensitive without being practically unreliable, if the shift is very small, and it can also appear non-sensitive without being demonstrably robust, if equivalence has not been established. 
Considering both perspectives allows us to distinguish metrics that merely show no detectable shift from metrics that provide positive evidence of stability under meaning-preserving rewriting.
We operationalize our work using the PeerPrism dataset~\cite{peerprism}, which comprises 4,044 LLM-generated rewrites derived from 674 human reviews of ICLR and NeurIPS submissions from 2021 to 2024. The rewritten reviews are designed to preserve the original review's judgments, arguments, and recommendations while altering wording and presentation. We evaluate 29 content-oriented metrics drawn from four prior works, ReviewEval~\cite{revieweval}, REMOR~\cite{remor}, RottenReviews~\cite{rottenreviews}, and ScholarPeer~\cite{scholarpeer}. We assess \emph{surface sensitivity} using paired significance testing and \emph{robustness} using equivalence testing, and we replicate the analysis with two different SOTA LLMs as judge models.

Under the primary analysis, among the 29 examined metrics, 23 assign significantly different scores to rewritten reviews whose evaluative content remains unchanged, while only 6 satisfy our robustness criterion. The finding is largely consistent across judge models, with 24 of 29 metrics receiving the same classification under both LLMs. The results suggest that many peer-review evaluation metrics partially conflate substantive review quality with linguistic presentation. Our findings suggest that robustness to meaning-preserving rewriting should be validated before LLM-as-a-judge metrics are used to compare human, AI-assisted, and AI-generated reviews. 

More concretely, this paper makes the following contributions: 
\textbf{(1)} 
 We formulate \emph{surface sensitivity} and \emph{robustness} as distinct but complementary criteria for evaluating peer-review metrics, separating statistically detectable score shifts from practically meaningful instability under meaning-preserving rewriting;
\textbf{(2)} We present a systematic analysis of \emph{surface sensitivity} and \emph{robustness} in LLM-as-a-judge metrics for peer-review evaluation, covering 29 content-oriented metrics drawn from four prior works using a dataset comprising 4,044 meaning-preserving rewrites of 674 human reviews from ICLR and NeurIPS; and \textbf{(3)} We show that sensitivity to surface-level rewriting is pervasive while robustness is comparatively rare, with results that are largely consistent across two different LLMs, highlighting the need to validate metric stability before drawing conclusions about human and AI-assisted review quality.

\section{Analytical Framework}
\label{sec:stats}

\begin{table*}[t]
    \vspace{-1em}
    \caption{Per-metric surface sensitivity and robustness using GPT-5-mini as the judge. Metrics are classified with Bonferroni-corrected Wilcoxon and TOST tests ($\alpha=0.05/29$). $\sigma_m$ is the standard deviation of paired differences, $\delta_m=0.2\sigma_m$ is the equivalence bound, and the final column reports whether the classification agrees with Gemini-2.5-Flash.}
    \vspace{-1em}
    \label{tab:sensitivity_tost_results}
    \centering
    \scriptsize
    \setlength{\tabcolsep}{2.5pt}
    \renewcommand{\arraystretch}{0.90}
    \begin{tabular}{@{}p{1.4cm}p{2.8cm}cp{0.55cm}p{0.55cm}rp{1.2cm}cp{1.0cm}cc@{}}
    \toprule
    \textbf{Source} &
    \textbf{Metric} &
    \textbf{Scale} &
    \boldmath$\sigma_m$ &
    \boldmath$\delta_m$ &
    \textbf{Mean $\Delta$} &
    \textbf{Wilcoxon $p$} &
    \textbf{Sensitive?} &
    \textbf{TOST $p$} &
    \textbf{Robust?} &
    \textbf{Gemini-2.5-Flash Agreement} \\
    \midrule
    
    \multirow{9}{*}{ReviewEval}
    & \cellcolor[HTML]{FFFDEB}Literature Comparison       & \cellcolor[HTML]{FFFDEB}[0,1]        & \cellcolor[HTML]{FFFDEB}0.332 & \cellcolor[HTML]{FFFDEB}0.066 & \cellcolor[HTML]{FFFDEB}+0.051 & \cellcolor[HTML]{FFFDEB}$8.68\!\times\!10^{-5}$  & \cellcolor[HTML]{FFFDEB}$\checkmark$ & \cellcolor[HTML]{FFFDEB}0.080                   & \cellcolor[HTML]{FFFDEB}--- & \cellcolor[HTML]{FFFDEB}$\checkmark$ \\
    & \cellcolor[HTML]{FFFDEB}Methodological Scrutiny     & \cellcolor[HTML]{FFFDEB}[0,1]        & \cellcolor[HTML]{FFFDEB}0.209 & \cellcolor[HTML]{FFFDEB}0.042 & \cellcolor[HTML]{FFFDEB}+0.060 & \cellcolor[HTML]{FFFDEB}$9.48\!\times\!10^{-17}$ & \cellcolor[HTML]{FFFDEB}$\checkmark$ & \cellcolor[HTML]{FFFDEB}0.996                   & \cellcolor[HTML]{FFFDEB}--- & \cellcolor[HTML]{FFFDEB}$\checkmark$ \\
    & \cellcolor[HTML]{FFFDEB}Results Interpretation      & \cellcolor[HTML]{FFFDEB}[0,1]        & \cellcolor[HTML]{FFFDEB}0.225 & \cellcolor[HTML]{FFFDEB}0.045 & \cellcolor[HTML]{FFFDEB}+0.082 & \cellcolor[HTML]{FFFDEB}$3.46\!\times\!10^{-23}$ & \cellcolor[HTML]{FFFDEB}$\checkmark$ & \cellcolor[HTML]{FFFDEB}1.000                   & \cellcolor[HTML]{FFFDEB}--- & \cellcolor[HTML]{FFFDEB}$\checkmark$ \\
    & \cellcolor[HTML]{FFFDEB}Theoretical Contributions   & \cellcolor[HTML]{FFFDEB}[0,1]        & \cellcolor[HTML]{FFFDEB}0.247 & \cellcolor[HTML]{FFFDEB}0.049 & \cellcolor[HTML]{FFFDEB}+0.044 & \cellcolor[HTML]{FFFDEB}$9.57\!\times\!10^{-9}$  & \cellcolor[HTML]{FFFDEB}$\checkmark$ & \cellcolor[HTML]{FFFDEB}0.238                   & \cellcolor[HTML]{FFFDEB}--- & \cellcolor[HTML]{FFFDEB}$\checkmark$ \\
    & \cellcolor[HTML]{FFFDEB}Logical Gaps Identification & \cellcolor[HTML]{FFFDEB}[0,1]        & \cellcolor[HTML]{FFFDEB}0.239 & \cellcolor[HTML]{FFFDEB}0.048 & \cellcolor[HTML]{FFFDEB}+0.053 & \cellcolor[HTML]{FFFDEB}$3.58\!\times\!10^{-12}$ & \cellcolor[HTML]{FFFDEB}$\checkmark$ & \cellcolor[HTML]{FFFDEB}0.756                   & \cellcolor[HTML]{FFFDEB}--- & \cellcolor[HTML]{FFFDEB}$\checkmark$ \\
    & \cellcolor[HTML]{FFFDEB}Overall Depth               & \cellcolor[HTML]{FFFDEB}[0,1]        & \cellcolor[HTML]{FFFDEB}0.149 & \cellcolor[HTML]{FFFDEB}0.030 & \cellcolor[HTML]{FFFDEB}+0.058 & \cellcolor[HTML]{FFFDEB}$4.56\!\times\!10^{-30}$ & \cellcolor[HTML]{FFFDEB}$\checkmark$ & \cellcolor[HTML]{FFFDEB}1.000                   & \cellcolor[HTML]{FFFDEB}--- & \cellcolor[HTML]{FFFDEB}$\checkmark$ \\
    & \cellcolor[HTML]{E9F4FE}Total Insights              & \cellcolor[HTML]{E9F4FE}[0,$\infty$) & \cellcolor[HTML]{E9F4FE}4.577 & \cellcolor[HTML]{E9F4FE}0.915 & \cellcolor[HTML]{E9F4FE}-0.356 & \cellcolor[HTML]{E9F4FE}0.002                    & \cellcolor[HTML]{E9F4FE}---          & \cellcolor[HTML]{E9F4FE}$7.92\!\times\!10^{-5}$ & \cellcolor[HTML]{E9F4FE}$\checkmark$ & \cellcolor[HTML]{E9F4FE}$\checkmark$ \\
    & \cellcolor[HTML]{E9F4FE}Actionable Insights         & \cellcolor[HTML]{E9F4FE}[0,$\infty$) & \cellcolor[HTML]{E9F4FE}4.938 & \cellcolor[HTML]{E9F4FE}0.988 & \cellcolor[HTML]{E9F4FE}-0.319 & \cellcolor[HTML]{E9F4FE}0.016                    & \cellcolor[HTML]{E9F4FE}---          & \cellcolor[HTML]{E9F4FE}$1.42\!\times\!10^{-5}$ & \cellcolor[HTML]{E9F4FE}$\checkmark$ & \cellcolor[HTML]{E9F4FE}$\checkmark$ \\
    & \cellcolor[HTML]{E9F4FE}Actionability Score         & \cellcolor[HTML]{E9F4FE}[0,1]        & \cellcolor[HTML]{E9F4FE}0.058 & \cellcolor[HTML]{E9F4FE}0.012 & \cellcolor[HTML]{E9F4FE}+0.001 & \cellcolor[HTML]{E9F4FE}0.909                    & \cellcolor[HTML]{E9F4FE}---          & \cellcolor[HTML]{E9F4FE}$1.92\!\times\!10^{-8}$ & \cellcolor[HTML]{E9F4FE}$\checkmark$ & \cellcolor[HTML]{E9F4FE}$\checkmark$ \\
    \midrule
    
    \multirow{7}{*}{REMOR}
    & \cellcolor[HTML]{E9F4FE}Criticism               & \cellcolor[HTML]{E9F4FE}[0,1] & \cellcolor[HTML]{E9F4FE}0.166 & \cellcolor[HTML]{E9F4FE}0.033 & \cellcolor[HTML]{E9F4FE}+0.017 & \cellcolor[HTML]{E9F4FE}0.002                    & \cellcolor[HTML]{E9F4FE}---          & \cellcolor[HTML]{E9F4FE}0.001                   & \cellcolor[HTML]{E9F4FE}$\checkmark$ & \cellcolor[HTML]{E9F4FE}--- \\
    & \cellcolor[HTML]{FFFDEB}Example                 & \cellcolor[HTML]{FFFDEB}[0,1] & \cellcolor[HTML]{FFFDEB}0.112 & \cellcolor[HTML]{FFFDEB}0.022 & \cellcolor[HTML]{FFFDEB}+0.018 & \cellcolor[HTML]{FFFDEB}$3.05\!\times\!10^{-5}$  & \cellcolor[HTML]{FFFDEB}$\checkmark$ & \cellcolor[HTML]{FFFDEB}0.110                    & \cellcolor[HTML]{FFFDEB}---          & \cellcolor[HTML]{FFFDEB}$\checkmark$ \\
    & \cellcolor[HTML]{FFFDEB}Importance \& Relevance & \cellcolor[HTML]{FFFDEB}[0,1] & \cellcolor[HTML]{FFFDEB}0.133 & \cellcolor[HTML]{FFFDEB}0.027 & \cellcolor[HTML]{FFFDEB}+0.048 & \cellcolor[HTML]{FFFDEB}$1.88\!\times\!10^{-27}$ & \cellcolor[HTML]{FFFDEB}$\checkmark$ & \cellcolor[HTML]{FFFDEB}1.000                    & \cellcolor[HTML]{FFFDEB}---          & \cellcolor[HTML]{FFFDEB}$\checkmark$ \\
    & \cellcolor[HTML]{E9F4FE}Materials \& Methods    & \cellcolor[HTML]{E9F4FE}[0,1] & \cellcolor[HTML]{E9F4FE}0.169 & \cellcolor[HTML]{E9F4FE}0.034 & \cellcolor[HTML]{E9F4FE}-0.005 & \cellcolor[HTML]{E9F4FE}0.336                    & \cellcolor[HTML]{E9F4FE}---          & \cellcolor[HTML]{E9F4FE}$5.98\!\times\!10^{-8}$ & \cellcolor[HTML]{E9F4FE}$\checkmark$ & \cellcolor[HTML]{E9F4FE}--- \\
    & \cellcolor[HTML]{FFFDEB}Praise                  & \cellcolor[HTML]{FFFDEB}[0,1] & \cellcolor[HTML]{FFFDEB}0.155 & \cellcolor[HTML]{FFFDEB}0.031 & \cellcolor[HTML]{FFFDEB}+0.032 & \cellcolor[HTML]{FFFDEB}$6.31\!\times\!10^{-16}$ & \cellcolor[HTML]{FFFDEB}$\checkmark$ & \cellcolor[HTML]{FFFDEB}0.552                    & \cellcolor[HTML]{FFFDEB}---          & \cellcolor[HTML]{FFFDEB}$\checkmark$ \\
    & \cellcolor[HTML]{E9F4FE}Results \& Discussion   & \cellcolor[HTML]{E9F4FE}[0,1] & \cellcolor[HTML]{E9F4FE}0.125 & \cellcolor[HTML]{E9F4FE}0.025 & \cellcolor[HTML]{E9F4FE}-0.006 & \cellcolor[HTML]{E9F4FE}0.146                    & \cellcolor[HTML]{E9F4FE}---          & \cellcolor[HTML]{E9F4FE}$1.30\!\times\!10^{-6}$ & \cellcolor[HTML]{E9F4FE}$\checkmark$ & \cellcolor[HTML]{E9F4FE}--- \\
    & \cellcolor[HTML]{FFFDEB}Suggestion \& Solution  & \cellcolor[HTML]{FFFDEB}[0,1] & \cellcolor[HTML]{FFFDEB}0.135 & \cellcolor[HTML]{FFFDEB}0.027 & \cellcolor[HTML]{FFFDEB}+0.028 & \cellcolor[HTML]{FFFDEB}$4.45\!\times\!10^{-10}$ & \cellcolor[HTML]{FFFDEB}$\checkmark$ & \cellcolor[HTML]{FFFDEB}0.557                    & \cellcolor[HTML]{FFFDEB}---          & \cellcolor[HTML]{FFFDEB}$\checkmark$ \\
    \midrule
    
    \multirow{8}{*}{RottenReviews}
    & \cellcolor[HTML]{FFFDEB}Comprehensiveness     & \cellcolor[HTML]{FFFDEB}[0,5]   & \cellcolor[HTML]{FFFDEB}0.305  & \cellcolor[HTML]{FFFDEB}0.061 & \cellcolor[HTML]{FFFDEB}+0.383 & \cellcolor[HTML]{FFFDEB}$4.97\!\times\!10^{-124}$ & \cellcolor[HTML]{FFFDEB}$\checkmark$ & \cellcolor[HTML]{FFFDEB}1.000 & \cellcolor[HTML]{FFFDEB}--- & \cellcolor[HTML]{FFFDEB}$\checkmark$ \\
    & \cellcolor[HTML]{FFFDEB}Objectivity           & \cellcolor[HTML]{FFFDEB}[0,5]   & \cellcolor[HTML]{FFFDEB}0.305  & \cellcolor[HTML]{FFFDEB}0.061 & \cellcolor[HTML]{FFFDEB}+0.372 & \cellcolor[HTML]{FFFDEB}$9.06\!\times\!10^{-120}$ & \cellcolor[HTML]{FFFDEB}$\checkmark$ & \cellcolor[HTML]{FFFDEB}1.000 & \cellcolor[HTML]{FFFDEB}--- & \cellcolor[HTML]{FFFDEB}$\checkmark$ \\
    & \cellcolor[HTML]{FFFDEB}Fairness              & \cellcolor[HTML]{FFFDEB}[0,5]   & \cellcolor[HTML]{FFFDEB}0.312  & \cellcolor[HTML]{FFFDEB}0.062 & \cellcolor[HTML]{FFFDEB}+0.310 & \cellcolor[HTML]{FFFDEB}$2.03\!\times\!10^{-102}$ & \cellcolor[HTML]{FFFDEB}$\checkmark$ & \cellcolor[HTML]{FFFDEB}1.000 & \cellcolor[HTML]{FFFDEB}--- & \cellcolor[HTML]{FFFDEB}$\checkmark$ \\
    & \cellcolor[HTML]{FFFDEB}Actionability         & \cellcolor[HTML]{FFFDEB}[0,5]   & \cellcolor[HTML]{FFFDEB}0.312  & \cellcolor[HTML]{FFFDEB}0.062 & \cellcolor[HTML]{FFFDEB}+0.291 & \cellcolor[HTML]{FFFDEB}$2.88\!\times\!10^{-94}$  & \cellcolor[HTML]{FFFDEB}$\checkmark$ & \cellcolor[HTML]{FFFDEB}1.000 & \cellcolor[HTML]{FFFDEB}--- & \cellcolor[HTML]{FFFDEB}$\checkmark$ \\
    & \cellcolor[HTML]{FFFDEB}Constructiveness      & \cellcolor[HTML]{FFFDEB}[0,5]   & \cellcolor[HTML]{FFFDEB}0.298  & \cellcolor[HTML]{FFFDEB}0.060 & \cellcolor[HTML]{FFFDEB}+0.311 & \cellcolor[HTML]{FFFDEB}$2.12\!\times\!10^{-105}$ & \cellcolor[HTML]{FFFDEB}$\checkmark$ & \cellcolor[HTML]{FFFDEB}1.000 & \cellcolor[HTML]{FFFDEB}--- & \cellcolor[HTML]{FFFDEB}$\checkmark$ \\
    & \cellcolor[HTML]{FFFDEB}Relevance Alignment   & \cellcolor[HTML]{FFFDEB}[0,5]   & \cellcolor[HTML]{FFFDEB}0.060  & \cellcolor[HTML]{FFFDEB}0.012 & \cellcolor[HTML]{FFFDEB}+0.012 & \cellcolor[HTML]{FFFDEB}$4.95\!\times\!10^{-8}$   & \cellcolor[HTML]{FFFDEB}$\checkmark$ & \cellcolor[HTML]{FFFDEB}0.446 & \cellcolor[HTML]{FFFDEB}--- & \cellcolor[HTML]{FFFDEB}$\checkmark$ \\
    & \cellcolor[HTML]{FFFDEB}Overall Quality       & \cellcolor[HTML]{FFFDEB}[0,100] & \cellcolor[HTML]{FFFDEB}12.435 & \cellcolor[HTML]{FFFDEB}2.487 & \cellcolor[HTML]{FFFDEB}+3.887 & \cellcolor[HTML]{FFFDEB}$9.84\!\times\!10^{-30}$  & \cellcolor[HTML]{FFFDEB}$\checkmark$ & \cellcolor[HTML]{FFFDEB}1.000 & \cellcolor[HTML]{FFFDEB}--- & \cellcolor[HTML]{FFFDEB}$\checkmark$ \\
    & \cellcolor[HTML]{FFFDEB}Overall Score         & \cellcolor[HTML]{FFFDEB}[0,100] & \cellcolor[HTML]{FFFDEB}3.085  & \cellcolor[HTML]{FFFDEB}0.617 & \cellcolor[HTML]{FFFDEB}+5.573 & \cellcolor[HTML]{FFFDEB}$1.68\!\times\!10^{-152}$ & \cellcolor[HTML]{FFFDEB}$\checkmark$ & \cellcolor[HTML]{FFFDEB}1.000 & \cellcolor[HTML]{FFFDEB}--- & \cellcolor[HTML]{FFFDEB}$\checkmark$ \\
    \midrule
    
    \multirow{5}{*}{ScholarPeer}
    & \cellcolor[HTML]{FFFDEB}Technical Accuracy      & \cellcolor[HTML]{FFFDEB}[1,10] & \cellcolor[HTML]{FFFDEB}0.881 & \cellcolor[HTML]{FFFDEB}0.176 & \cellcolor[HTML]{FFFDEB}-0.419 & \cellcolor[HTML]{FFFDEB}$5.97\!\times\!10^{-161}$ & \cellcolor[HTML]{FFFDEB}$\checkmark$ & \cellcolor[HTML]{FFFDEB}1.000 & \cellcolor[HTML]{FFFDEB}--- & \cellcolor[HTML]{FFFDEB}$\checkmark$ \\
    & \cellcolor[HTML]{FFFDEB}Constructive Value      & \cellcolor[HTML]{FFFDEB}[1,10] & \cellcolor[HTML]{FFFDEB}0.891 & \cellcolor[HTML]{FFFDEB}0.178 & \cellcolor[HTML]{FFFDEB}-0.707 & \cellcolor[HTML]{FFFDEB}$<\!10^{-300}$            & \cellcolor[HTML]{FFFDEB}$\checkmark$ & \cellcolor[HTML]{FFFDEB}1.000 & \cellcolor[HTML]{FFFDEB}--- & \cellcolor[HTML]{FFFDEB}--- \\
    & \cellcolor[HTML]{FFFDEB}Analytical Depth        & \cellcolor[HTML]{FFFDEB}[1,10] & \cellcolor[HTML]{FFFDEB}0.782 & \cellcolor[HTML]{FFFDEB}0.156 & \cellcolor[HTML]{FFFDEB}-1.192 & \cellcolor[HTML]{FFFDEB}$<\!10^{-300}$            & \cellcolor[HTML]{FFFDEB}$\checkmark$ & \cellcolor[HTML]{FFFDEB}1.000 & \cellcolor[HTML]{FFFDEB}--- & \cellcolor[HTML]{FFFDEB}$\checkmark$ \\
    & \cellcolor[HTML]{FFFDEB}Novelty \& Significance & \cellcolor[HTML]{FFFDEB}[1,10] & \cellcolor[HTML]{FFFDEB}1.203 & \cellcolor[HTML]{FFFDEB}0.241 & \cellcolor[HTML]{FFFDEB}-0.265 & \cellcolor[HTML]{FFFDEB}$3.44\!\times\!10^{-41}$  & \cellcolor[HTML]{FFFDEB}$\checkmark$ & \cellcolor[HTML]{FFFDEB}0.894 & \cellcolor[HTML]{FFFDEB}--- & \cellcolor[HTML]{FFFDEB}$\checkmark$ \\
    & \cellcolor[HTML]{FFFDEB}Overall                 & \cellcolor[HTML]{FFFDEB}[1,10] & \cellcolor[HTML]{FFFDEB}0.737 & \cellcolor[HTML]{FFFDEB}0.147 & \cellcolor[HTML]{FFFDEB}-0.595 & \cellcolor[HTML]{FFFDEB}$<\!10^{-300}$            & \cellcolor[HTML]{FFFDEB}$\checkmark$ & \cellcolor[HTML]{FFFDEB}1.000 & \cellcolor[HTML]{FFFDEB}--- & \cellcolor[HTML]{FFFDEB}--- \\
    \bottomrule
    \end{tabular}
    \vspace{-1em}
    \end{table*}

Our objective is to evaluate whether peer review evaluation metrics measure the substantive assessment expressed in a review or also respond to its surface form. We consider pairs of reviews in which an original and its meaning-preserving rewrite express the same evaluative content but differ in wording, phrasing, organization, or style. A reliable content-oriented metric should remain stable under such transformations, since the quality of the underlying review as an evaluation has not changed.

Let $R$ denote an original peer review and $R'$ a meaning-preserving rewrite that preserves the evaluative content of $R$ while modifying only its surface realization. A metric designed to measure properties such as analytical depth, constructiveness, or aspect coverage should assign similar scores to $R$ and $R'$, since the underlying assessment has not changed.

This setting motivates two complementary notions of metric reliability. \emph{Surface sensitivity} asks whether meaning-preserving rewriting induces a systematic change in metric scores. \emph{Robustness} asks whether any such change is small enough to be practically negligible. Surface sensitivity identifies systematic score shifts, whereas robustness provides positive evidence that their magnitude remains within an acceptable range.

For each metric $m$ and paired instance $(R_i, R'_i)$, we 
compute the paired score difference
$
d_i = S_m(R'_i) - S_m(R_i),
$
\noindent where $S_m(\cdot)$ denotes the score assigned by 
metric $m$. 

Let $D_m = \{d_i^{(m)}\}_{i=1}^{n}$ denote the 
collection of paired differences for metric $m$, which forms 
the basis of both analyses. To assess surface sensitivity, we 
test whether rewriting induces a systematic shift using the 
Wilcoxon signed-rank test~\cite{wilcoxon1992individual}:
\vspace{-0.3em}
\begin{equation}
H_0:\; \operatorname{median}(D_m) = 0
\qquad\text{vs.}\qquad
H_1:\; \operatorname{median}(D_m) \neq 0.
\end{equation}

Rejecting $H_0$ indicates that the metric assigns systematically different scores to original and rewritten reviews. Because the rewrite is intended to preserve evaluative content, such a shift is interpreted as evidence that the metric is affected by surface-level presentation. We assess surface sensitivity using the Wilcoxon signed-rank test~\cite{wilcoxon1992individual}, a non-parametric procedure for paired comparisons that does not require normally distributed differences. 

To assess \emph{robustness}, we use the Two One-Sided Tests (TOST) procedure~\cite{lakens2017equivalence}. Whereas the Wilcoxon test asks whether a non-zero shift can be detected, equivalence testing asks whether the shift is sufficiently small to be considered practically negligible. For each metric $m$, we define an equivalence bound $\delta_m > 0$ and test:
\vspace{-0.5em}
\begin{equation}
H_0:\; |\mu_m| \ge \delta_m
\qquad\text{vs.}\qquad
H_1:\; |\mu_m| < \delta_m,
\end{equation}

\noindent where $\mu_m = \mathbb{E}[D_m]$ is the mean paired difference. A metric is considered robust if the TOST procedure rejects the null hypothesis, providing positive evidence that the true mean shift lies within the equivalence interval $[-\delta_m,+\delta_m]$. Following the conventional small-effect threshold of $d=0.2$~\cite{cohen1992statistical} and recommendations for equivalence testing when no external criterion for practical significance is available~\cite{lakens2017equivalence}, we set $\delta_m = 0.2\sigma_m,$ 
where $\sigma_m$ is the standard deviation of the paired differences for metric $m$. This choice operationalizes practical equivalence as a small standardized change, thereby providing a comparable criterion across metrics with different scoring scales. Because the equivalence margin is determined by the observed variability of each metric, the resulting robustness classification is conditional on this criterion and should not be interpreted as evidence of invariance to rewriting under all possible equivalence thresholds.

Both tests are applied independently to all evaluated metrics. To control the family-wise error rate, we apply Bonferroni correction~\cite{bonferroni1936teoria}
$\alpha_{\text{corrected}}
=
\frac{0.05}{M},$ 
where $M$ is the number of evaluated metrics. The same corrected threshold is used for sensitivity and robustness testing. Combining the Wilcoxon and TOST outcomes allows us to distinguish four forms of metric behavior. A metric is classified as \textbf{sensitive only} when the Wilcoxon test rejects the null hypothesis, but the TOST procedure does not, meaning that rewriting induces a systematic score shift whose magnitude cannot be treated as practically negligible. A metric is classified as \textbf{robust only} when the Wilcoxon test does not reject the null hypothesis and the TOST procedure does, meaning that no systematic shift is detected and equivalence is positively established. A metric falls into the \textbf{sensitive $\cap$ robust} category when both tests reject their null hypotheses, indicating that rewriting produces a statistically detectable shift, but one whose magnitude remains within the equivalence bound. Finally, a metric is classified as \textbf{inconclusive} when neither test rejects its null hypothesis, meaning that no systematic shift is detected, but equivalence is also not established.

\section{Experimental Setup}
\label{sec:experimental_setup}

 \textbf{Peer Review Evaluation Metrics.} We evaluate metrics introduced in four prior works: ReviewEval~\cite{revieweval}, REMOR~\cite{remor}, RottenReviews~\cite{rottenreviews}, and ScholarPeer~\cite{scholarpeer}. Together, these metrics cover analytical depth, actionability, review-component coverage, holistic quality, and human-anchored comparative judgment. Metrics from ReviewEval focus on depth and actionability, REMOR measures explicit review components, RottenReviews evaluates broad quality dimensions such as fairness and constructiveness, and ScholarPeer compares reviews against a paired human-written review, allowing us to examine whether human anchoring affects sensitivity to rewriting. Because our goal is to test whether substantive metrics are affected by surface-level variation, we retain only content-oriented metrics, including those for analytical depth, constructiveness, actionability, relevance, and aspect coverage, as reported in Table~\ref{tab:sensitivity_tost_results}. We exclude metrics that explicitly target style or presentation, namely \textit{Presentation \& Reporting} from REMOR and \textit{Usage of Technical Terms} and \textit{Clarity \& Readability} from RottenReviews. The resulting analysis contains $M=29$ metrics, giving $\alpha_{\text{corrected}} = 0.05/29 \approx 0.00172$.

 \textbf{Peer Review Dataset.} We use PeerPrism~\cite{peerprism}, a benchmark designed to separate evaluative content from surface realization in peer reviews. PeerPrism contains 160 ICLR and NeurIPS papers from 2021--2024. Our analysis uses only the \textit{rewritten} partition, which contains 4,044 faithful rewrites of 674 original human reviews. These rewrites preserve each review's judgments, arguments, strengths, weaknesses, and recommendations while changing wording, phrasing, and style. The analysis represents all papers and rewriting conditions. ReviewEval depth/actionability metrics and REMOR use one review per paper and condition, while RottenReviews aggregates scores across all available reviews at the paper level. The data have a nested structure, as multiple rewritten variants originate from the same underlying reviews and multiple reviews may correspond to the same paper. Although the metric frameworks differ in their evaluation granularity, this dependence structure should be considered when interpreting individual metric-level statistical results. We therefore emphasize the broader patterns observed across metrics.

 \textbf{Human Validation of Rewrites.} Our framework assumes that $R'$ preserves the evaluative content of $R$. To validate this, we conducted a human annotation study on 50 randomly sampled original--rewrite pairs. Three computer science graduate students annotated the pairs, with each pair evaluated by two annotators. Annotators rated how faithfully the rewritten review preserved the original review's content on an overall assessment on a 1--5 Likert scale. Additionally they indicated whether the original and rewritten reviews would provide essentially equivalent feedback to a paper author using a binary Yes or No judgment.

Faithfulness scores were high across annotators, with mean ratings of 4.28, 4.96, and 4.28 out of 5. Agreement was also strong: 90\% within-one-point agreement for faithfulness, 92\% exact agreement for binary equivalence, and 94\% of annotations indicating equivalent feedback. Overall, this validation supports the content faithfulness of the rewrites and their use for testing surface sensitivity and robustness.  Detailed annotation guidelines are provided in the project repository.

 \textbf{Judge Models and Scoring Protocol.} All evaluations are performed using GPT-5-mini~\cite{singh2025openai} as the judge model. For each metric, we use the original prompt and scoring rubric from the work that introduced the metric without modification. To assess whether the findings depend on the judge model, we repeat the full analysis using Gemini-2.5-Flash and compare the resulting sensitivity and robustness classifications against the GPT-5-mini results.

\section{Results and Findings}

\begin{figure}[t]
 \vspace{-0em}
    \includegraphics[width=0.9\linewidth]{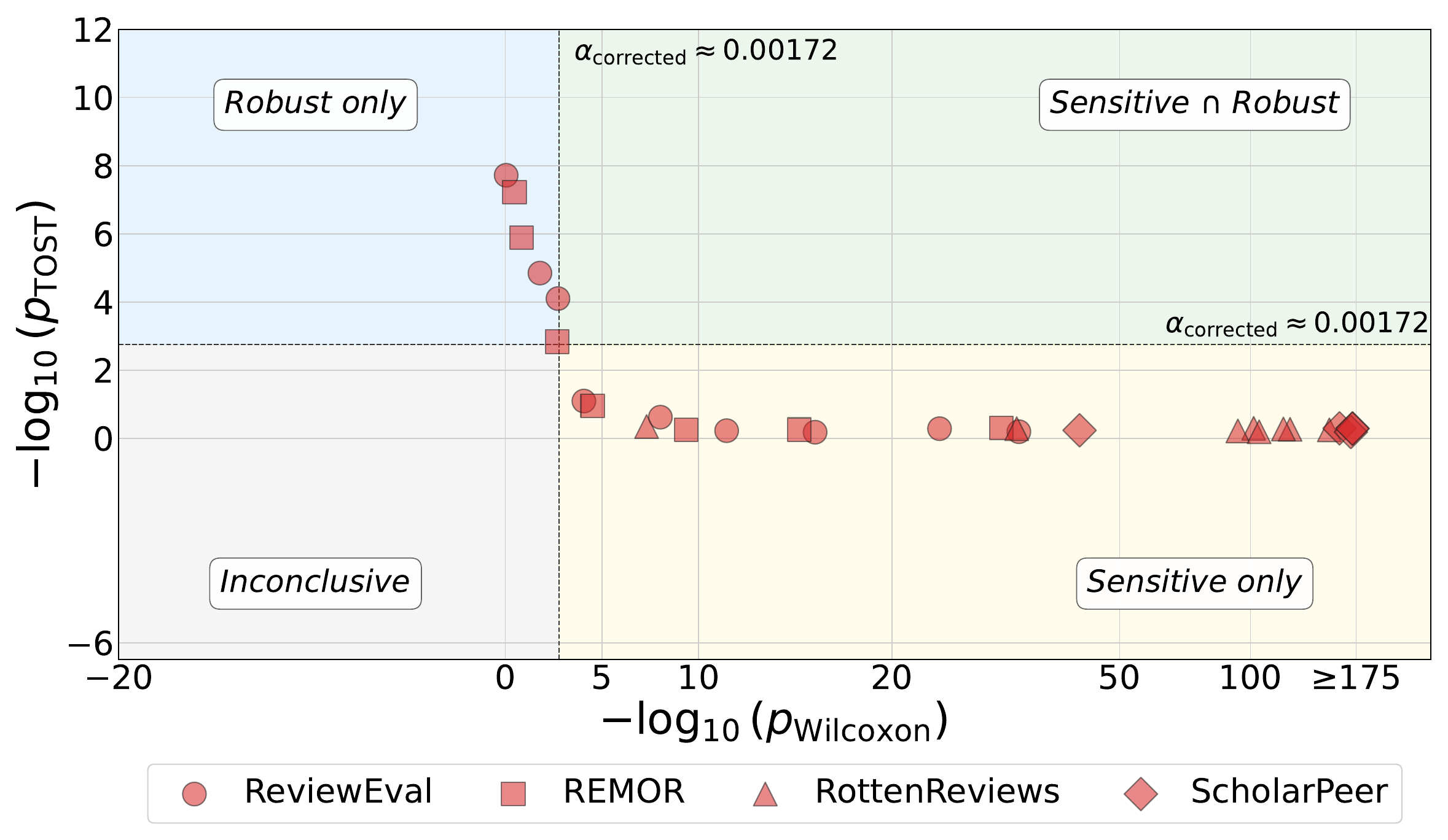}
    \caption{Sensitivity and robustness evidence for all metrics.
The four shaded areas correspond to 
\textit{Robust only}, \textit{Sensitive only},
\textit{Sensitive $\cap$ Robust}, and \textit{Inconclusive} regions.
}
    \label{fig:p_val_summary}  
\end{figure}

Table~\ref{tab:sensitivity_tost_results} reports the per-metric results, and Figure~\ref{fig:p_val_summary} summarizes the joint evidence from the Wilcoxon and TOST procedures. 
The question is whether metrics intended to evaluate substantive review quality remain stable when the same content is expressed in a different linguistic form. 
The overall observed pattern is that many LLM-as-a-judge metrics respond significantly to how that evaluation is written. More detailed observations are as follows:

 \textbf{(1) Surface sensitivity is a broad measurement problem.}
Rewriting reviews while preserving their content still causes many peer-review evaluation metrics to assign systematically different scores. This is shown in Table~\ref{tab:sensitivity_tost_results} by the \textit{Sensitive?} column and the Bonferroni-significant Wilcoxon $p$-values. Since the rewrites preserve the original review content, these shifts cannot be interpreted as ordinary differences in review quality; they indicate sensitivity to linguistic realization. This is especially concerning for metrics that require broad interpretive judgment, such as analytical depth, comprehensiveness, objectivity, fairness, constructiveness, and overall quality. Such judgments may be affected by fluency, organization, explicitness, and tone, even when the underlying critique is unchanged. The pattern appears across metrics: ReviewEval's depth-oriented metrics, RottenReviews' broad quality metrics, and ScholarPeer's human-anchored dimensions are especially sensitive, while several REMOR dimensions also show sensitivity. This suggests that surface sensitivity is not an artifact of a single metric source or scoring scale, but a broader challenge in LLM-as-a-judge peer-review evaluation.

\textbf{(2) Reliable metrics tend to be anchored in concrete review content.}
The metrics that appear reliable under the robustness criterion share an important property. They evaluate relatively concrete, localized, or explicitly identifiable aspects of a review rather than broad qualitative impressions. Under the TOST criterion, the metrics that satisfy our robustness criterion are ReviewEval's \textit{Total Insights}, \textit{Actionable Insights}, and \textit{Actionability Score}, and REMOR's \textit{Criticism}, \textit{Materials \& Methods}, and \textit{Results \& Discussion}. These metrics are less tied to overall polish and more tied to identifiable evaluative content. For metric design, this suggests that concrete content-based judgments are more stable under rewriting than global quality assessments. This does not mean component-level metrics are sufficient, since high-quality reviewing also requires synthesis and prioritization, but it suggests that broad constructs may need to be decomposed into explicit subproperties or validated against meaning-preserving rewrites.

\textbf{(3) Sensitivity and robustness identify different kinds of reliability.}
The joint view in Figure~\ref{fig:p_val_summary} shows why the Wilcoxon and TOST tests should be interpreted together. A metric can show a statistically detectable shift under rewriting while still remaining practically stable if the shift is small. Conversely, a metric can fail to show a significant shift without providing positive evidence that the effect is negligible. The joint classification therefore separates metrics that merely lack evidence of sensitivity from metrics that provide affirmative evidence of robustness. Most metrics fall into the \textit{Sensitive only} region under our primary analysis, meaning that rewriting induces statistically detectable score shifts for which equivalence is not established under the chosen bound. These results provide evidence of instability under surface-level rewriting, although the exact metric-level classifications depend on the statistical assumptions and equivalence criterion. By contrast, \textit{Total Insights}, \textit{Actionable Insights}, \textit{Actionability Score}, \textit{Criticism}, \textit{Materials \& Methods}, and \textit{Results \& Discussion} fall into the \textit{Robust only} region. Reliability also varies across metric sources. ReviewEval includes both sensitive depth metrics and robust insight-based metrics, REMOR contains several robust component-level metrics, RottenReviews is largely sensitive, and ScholarPeer has no metric that satisfies the robustness criterion.

\textbf{(4) Directionality reveals metric-specific presentation effects.}
The Mean $\Delta$ column in Table~\ref{tab:sensitivity_tost_results} shows that rewriting affects different metric sources in different directions. For ReviewEval, REMOR, and RottenReviews, rewritten reviews generally receive higher scores than the original human reviews, with especially large positive shifts for RottenReviews metrics such as \textit{Comprehensiveness} ($+0.383$) and \textit{Overall Score} ($+5.573$). This suggests that LLM judges often treat improved fluency, organization, or explicitness as evidence of stronger review quality. ScholarPeer shows the opposite pattern as its metrics assign lower scores to rewritten reviews, even though the evaluative content is preserved.
A plausible explanation is its human-anchored scoring design, where reviews are evaluated relative to a paired human-written reference. Under this setup, rewriting may change the perceived relationship between the rewritten review and the human reference, reducing the score even when the underlying evaluation remains aligned. This indicates that human anchoring does not eliminate surface sensitivity. It may instead introduce a different presentation-dependent behavior in which the metric becomes sensitive to perceived similarity to the reference rather than to evaluative substance alone.

\textbf{(5) The findings are not judge-model artifacts.}
The replication with Gemini-2.5-Flash indicates that the main conclusions are not specific to GPT-5-mini. As shown in the right-most column of Table~\ref{tab:sensitivity_tost_results}, under our primary analysis, most metric classifications remain unchanged across judge models, and the few disagreements occur close to the statistical decision boundary. This consistency provides evidence that the observed behavior is not solely an idiosyncrasy of a single judge model. Replacing one strong judge model with another is therefore unlikely to resolve the issue on its own. The more fundamental problem lies in how peer-review quality metrics are specified, how much they rely on holistic judgment, and whether they are validated against meaning-preserving changes in surface form.

\section{Conclusion}
We presented a robustness analysis of LLM-as-a-judge metrics for peer-review evaluation using meaning-preserving rewrites. Many metrics intended to assess substantive review quality were sensitive to changes in wording and presentation despite preserved evaluative content, while robust metrics tended to focus on concrete review content. These findings suggest that robustness to meaning-preserving rewriting should be validated before such metrics are used to compare human-written, AI-assisted, and AI-generated reviews.

\section*{GenAI Usage Disclosure}
Large language models were used in the preparation of this manuscript for 
grammar correction, prose polishing, and coding support in the implementation 
of the experimental pipeline. All research design, analysis, interpretation 
of results, and scholarly conclusions are the work of the authors.


\bibliographystyle{ACM-Reference-Format}
\balance
\bibliography{references-base}


\end{document}